\documentclass[11pt,a4paper]{article}
\usepackage[hyperref]{conll2017}
\usepackage{times}
\usepackage{latexsym}

\usepackage[english]{babel}
\usepackage[T1]{fontenc}
\usepackage[utf8]{inputenc}

\usepackage{url}
\usepackage{graphicx}
\usepackage{multirow}
\usepackage{amsthm}
\usepackage{amssymb}
\usepackage{amsmath}
\usepackage{amsfonts}
\usepackage{color}
\usepackage{mdwlist}
\usepackage{microtype}
\usepackage{tabularx}
\usepackage{booktabs}
\usepackage{color,colortbl}
\usepackage[ruled,vlined, linesnumbered]{algorithm2e}
\usepackage[normalem]{ulem}
\usepackage{xcolor}
\usepackage{subcaption}

\definecolor{Gray}{gray}{0.9}

\newcolumntype{b}{X}
\newcolumntype{m}{>{\hsize=.6\hsize}X}
\newcolumntype{s}{>{\hsize=.33\hsize}X}

\makeatletter
\newcommand{\removelatexerror}{\let\@latex@error\@gobble}
\makeatother

\aclfinalcopy 

\newcommand{\com}[1]{}

\title{Automatic Selection of Context Configurations for Improved Class-Specific Word Representations}

\author{Ivan Vuli\'c$^{\mathbf{1}}$, ~ Roy Schwartz$^{\mathbf{2, 3}}$, ~ Ari Rappoport$^{\mathbf{4}}$  \\
\bf{Roi Reichart}$^{\mathbf{5}}$, ~ {Anna Korhonen}$^{\mathbf{1}}$\\
$^{\mathbf{1}}$ Language Technology Lab, DTAL, University of Cambridge \\
$^{\mathbf{2}}$ CS \& Engineering, University of Washington and $^{\mathbf{3}}$Allen Institute for AI\\
$^{\mathbf{4}}$ Institute of Computer Science, The Hebrew University of Jerusalem\\
$^{\mathbf{5}}$ Faculty of Industrial Engineering and Management, Technion, IIT \\
\texttt{\{iv250,alk23\}@cam.ac.uk} \hspace{0.8em} \texttt{roysch@cs.washington.edu} \\ \texttt{arir@cs.huji.ac.il} \hspace{0.8em}\texttt{roiri@ie.technion.ac.il}}

\date{}

\begin{document}
\maketitle
\begin{abstract}

This paper is concerned with identifying contexts useful for training word representation models for different word classes such as adjectives (A), verbs (V), and nouns (N). We introduce a simple yet effective framework for an automatic selection of {\em class-specific context configurations}. We construct a context configuration space based on universal dependency relations between words, and efficiently search this space with an adapted beam search algorithm. In word similarity tasks for each word class, we show that our framework is both effective and efficient. Particularly, it improves the Spearman's $\rho$ correlation with human scores on SimLex-999 over the best previously proposed class-specific contexts by 6 (A), 6 (V) and 5 (N) $\rho$ points. With our selected context configurations, we train on only 14\% (A), 26.2\% (V), and 33.6\% (N) of all dependency-based contexts, resulting in a reduced training time. Our results generalise: we show that the configurations our algorithm learns for one English training setup outperform previously proposed context types in another training setup for English. Moreover, basing the configuration space on universal dependencies, it is possible to \textit{transfer} the learned configurations to German and Italian. We also demonstrate improved per-class results over other context types in these two languages.


\end{abstract}

\section{Introduction}
\label{s:intro}
Dense real-valued word representations (embeddings) have become ubiquitous in NLP, serving as invaluable features in a broad range of tasks \cite{Turian:2010acl,Collobert:2011jmlr,Chen:2014emnlp}. The omnipresent \texttt{word2vec} skip-gram model with negative sampling (SGNS) \cite{Mikolov:2013nips} is still considered a robust and effective choice for a word representation model, due to its simplicity, fast training, as well as its solid performance across semantic tasks \cite{Baroni:2014acl,Levy:2015tacl}. The original SGNS implementation learns word representations from local bag-of-words contexts (BOW). However, the underlying  model is equally applicable with other context types \cite{Levy:2014acl}.

Recent work suggests that ``not all contexts are created equal''. For example, reaching beyond standard BOW contexts towards contexts based on dependency parses \cite{Bansal:2014acl,Melamud:2016naacl} or symmetric patterns \cite{Schwartz:2015conll,Schwartz:2016naacl} yields significant improvements in learning representations for particular word classes such as {\em adjectives (A)} and {\em verbs (V)}. Moreover, \newcite{Schwartz:2016naacl} demonstrated 
that a subset of dependency-based contexts which covers only coordination structures is particularly effective for SGNS training, both in terms of the quality 
of the induced representations and in the reduced training time of the model.
Interestingly, they also demonstrated that despite the success with adjectives and verbs, BOW contexts are still the optimal choice when learning representations 
for {\em nouns (N)}. 

In this work, we propose a simple yet effective framework for selecting {\em  context configurations}, 
which yields improved representations for verbs, adjectives, {\em and} nouns. We start with a definition of our context 
configuration space (Sect. \ref{ss:space}). Our basic definition of a context 
refers to a single typed (or labeled) dependency link between words (e.g., the \texttt{amod} link or the \texttt{dobj} link).
Our configuration space then naturally consists of all  possible subsets of the set of labeled dependency links between words. We employ the universal dependencies (UD) scheme to make our framework applicable across languages. 
We then describe (Sect. \ref{ss:pools}) our adapted beam search algorithm that aims to select an optimal context configuration for a given word class. 

We show that SGNS requires different context configurations to produce improved results for each word class. 
For instance, our algorithm detects that the combination of \texttt{amod} and \texttt{conj} contexts 
is effective for adjective representation. Moreover, 
some contexts that boost representation learning for one word class (e.g., \texttt{amod} contexts for adjectives) 
may be uninformative when learning representations for another class (e.g., \texttt{amod} for verbs). By removing such dispensable contexts, we are able both to speed up the SGNS training and to improve representation quality. 

We first experiment with the task of predicting similarity scores for the A/V/N portions of the benchmarking 
SimLex-999 evaluation set, running our algorithm in a standard SGNS experimental setup \cite{Levy:2015tacl}. 
When training SGNS with our learned context configurations it outperforms SGNS trained with the best previously proposed 
context type {\em for each word class}: the improvements in Spearman's $\rho$ rank correlations are 6 (A), 6 (V), and 5 (N) points. We also show that by building context configurations we obtain improvements on the entire SimLex-999 (4 $\rho$ points over the best baseline). Interestingly, this context configuration is not the optimal configuration for any word class.


We then demonstrate that our approach is robust by showing that transferring the optimal configurations learned in the above setup to three other setups yields improved performance. First, the above context configurations, learned with the SGNS training on the English Wikipedia corpus, have an even stronger impact on SimLex999 performance when SGNS is trained on a larger corpus. Second, the transferred configurations also result in competitive performance on the task of solving class-specific TOEFL questions. Finally, we transfer the learned context configurations across languages: these configurations improve the SGNS performance when trained with German or Italian corpora and evaluated on 
class-specific subsets of the multilingual SimLex-999 \cite{Leviant:2015arxiv}, without any language-specific tuning.

\vspace{-0.0em}
\section{Related Work}
\label{s:rw}
Word representation models typically train on  ({\it word, context}) pairs.
Traditionally, most models use bag-of-words (BOW) contexts, which represent a word using its neighbouring words, irrespective of the syntactic or semantic relations between them \cite[inter alia]{Collobert:2011jmlr,Mikolov:2013nips,Mnih:2013nips,Pennington:2014emnlp}. Several alternative context types have been proposed, motivated by the limitations of BOW contexts, most notably their focus on topical rather than functional similarity (e.g., \textit{coffee:cup} vs. \textit{coffee:tea}). These include dependency contexts \cite{Pado:2007cl,Levy:2014acl}, pattern contexts \cite{Baroni:2010cogsci,Schwartz:2015conll} and substitute vectors \cite{Yatbaz:2012emnlp,Melamud:2015naacl}.

Several recent studies examined the effect of context types on word representation learning. 
\newcite{Melamud:2016naacl} compared three context types on a set of intrinsic and extrinsic evaluation setups: BOW, dependency links, and substitute vectors. They show that the optimal type largely depends on the task at hand, with dependency-based contexts displaying strong performance on semantic similarity tasks. \newcite{Vulic:2016acluniversal} extended the comparison to more languages, reaching similar conclusions. \newcite{Schwartz:2016naacl}, showed that symmetric patterns are useful as contexts for V and A similarity, while BOW still works best for nouns. They also indicated that coordination structures, a particular dependency link, are more useful for verbs and adjectives than the entire set of dependencies. In this work, we generalise their approach: our algorithm systematically and efficiently searches the space of dependency-based context configurations, yielding {\em class-specific} representations with substantial gains \textit{for all three word classes}.

Previous attempts on specialising word representations for a particular relation (e.g., similarity vs relatedness, antonyms) operate in one of two frameworks: (1) modifying the prior or the regularisation of the original training procedure \cite{Yu:2014acl,Wieting:2015tacl,Liu:2015acl,Kiela:2015emnlpemb,Ling:2015emnlp}; (2) post-processing procedures which use lexical knowledge to refine previously trained word vectors \cite{Faruqui:2015naacl,Wieting:2015tacl,Mrksic:2017ar}. Our work suggests that the induced representations can be specialised by directly training the word representation model with carefully selected contexts.

\section{Context Selection: Methodology}
\label{s:metho}
\begin{figure}[t]
\centering
\includegraphics[width=0.95\linewidth]{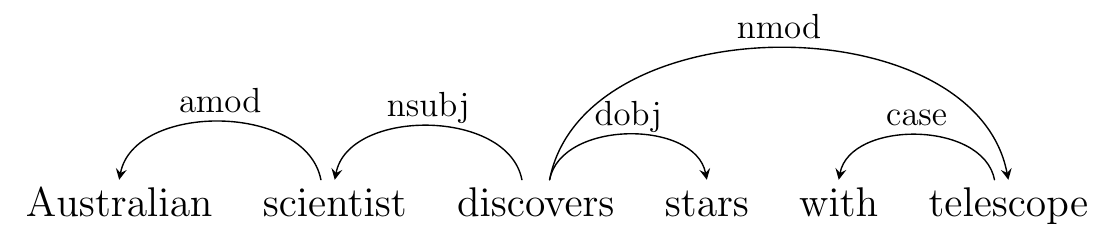}
\vspace{-0.2em}
\includegraphics[width=0.95\linewidth]{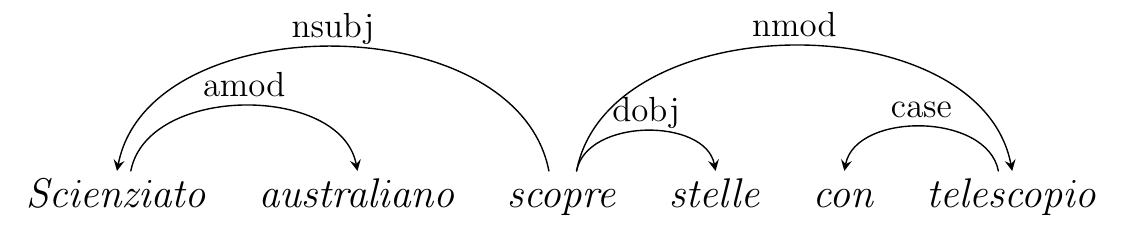}
\vspace{-0.2em}
\includegraphics[width=0.95\linewidth]{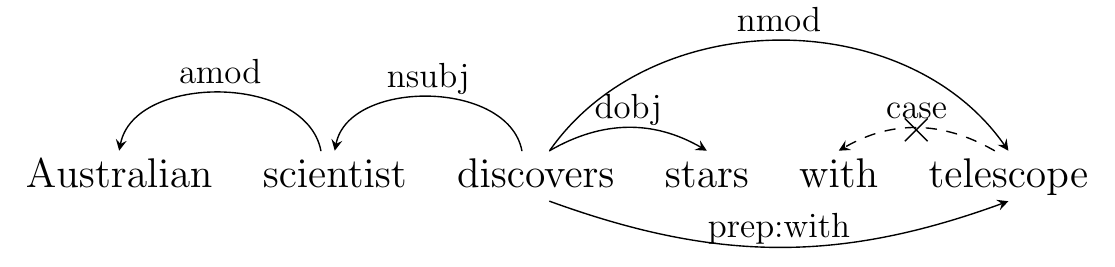}
\vspace{-0.0em}
\caption{Extracting dependency-based contexts. \textbf{Top}: An example English sentence from \cite{Levy:2014acl}, now UD-parsed. \textbf{Middle}: the same sentence in Italian, UD-parsed. Note the similarity between the two parses which suggests that our context selection framework may be extended to other languages. 
\textbf{Bottom}: prepositional arc collapsing. The uninformative short-range \texttt{case} arc is removed, while a ``pseudo-arc'' specifying the exact link (\texttt{prep:with}) between {\em discovers} and {\em telescope} is added.}
\vspace{-0.4em}
\label{fig:rex}
\end{figure}

The goal of our work is to develop a methodology for the identification of optimal context configurations for  
word representation model training. We hope to get improved word representations and, at the same time, cut down the training time of the word representation model. Fundamentally, we are not trying to design a new word representation model, but rather to find valuable configurations for existing algorithms. 

The motivation to search for such training context configurations lies in the intuition that the distributional hypothesis \cite{Harris:1954} 
should not necessarily be made with respect to BOW contexts. Instead, it may be restated as a series of statements according to particular word relations. For example, the hypothesis can be restated as: ``two adjectives are similar if they modify similar nouns'', which is captured by the \texttt{amod} typed dependency relation. This could also be reversed to reflect noun similarity by saying that ``two nouns are similar if they are modified by similar adjectives''. In another example, ``two verbs are similar if they are used as predicates of similar nominal subjects'' (the \texttt{nsubj} and \texttt{nsubjpass} dependency relations).

First, we have to define an expressive context configuration space that contains potential training configurations and is effectively decomposed so that useful configurations may be sought algorithmically. We can then continue by designing a search algorithm over the configuration space. 

\subsection{Context Configuration Space}
\label{ss:space}

We focus on the configuration space based on dependency-based contexts (DEPS) \cite{Pado:2007cl,Utt:2014tacl}. We choose this space due to multiple reasons. First, dependency structures are known to be very useful in capturing functional relations between words, even if these relations are long distance. 
Second, they have been proven useful in learning word embeddings \cite{Levy:2014acl,Melamud:2016naacl}. Finally, owing to the recent development of the Universal Dependencies (UD) annotation scheme \cite{McDonald:2013acl,Nivre:2015ud}\footnote{http://universaldependencies.org/ (V1.4 used)}
it is possible to reason over dependency structures in a multilingual manner (e.g., Fig.~\ref{fig:rex}). Consequently, a search algorithm in such DEPS-based configuration space can be developed for multiple languages based on the same design principles. Indeed, in this work we show that the optimal configurations for English translate to improved representations in two additional languages, German and Italian.

And so, given a (UD-)parsed training corpus, for each target word $w$ with modifiers $m_1,\ldots,m_k$ and a head $h$, the word $w$ is paired with context elements $m_1\_r_1,\ldots,m_k\_r_k,h\_r_h^{-1}$, where $r$ is the type of the dependency relation between the head and the modifier (e.g., \texttt{amod}), and $r^{-1}$ denotes an inverse relation. To simplify the presentation, we adopt the assumption that all training data for the word representation model are in the form of such $(word, context)$ pairs \cite{Levy:2014acl,Levy:2014nips}, where $word$ is the current target word, and $context$ is its observed context (e.g., BOW, positional, dependency-based). A naive version of DEPS extracts contexts from the parsed corpus without any post-processing. Given the example from Fig.~\ref{fig:rex}, the DEPS contexts of {\em discovers} are: {\em scientist\_nsubj}, {\em stars\_dobj}, {\em telescope\_nmod}.

DEPS not only emphasises functional similarity, but also provides a natural implicit grouping of related contexts. For instance, all pairs with the shared relation $r$ and $r^{-1}$ are taken as an $r$-based {\em context bag}, e.g., the pairs $\{${\em (scientist, Australian\_amod)}, {\em (Australian, scientist\_$amod^{-1}$)}$\}$ from Fig.~\ref{fig:rex} are inserted into the \texttt{amod} context bag, while $\{${\em (discovers, stars\_dobj)}, {\em (stars, discovers\_$dobj^{-1}$)}$\}$ are labelled with \texttt{dobj}.

Assume that we have obtained $M$ distinct dependency relations $r_1,\ldots,r_M$ after parsing and post-processing the corpus. The $j$-th {\em individual context bag}, $j=1,\ldots,M$, labelled $r_j$, is a bag (or a multiset) of $(word, context)$ pairs where $context$ has one of the following forms: $v\_r_j$ or $v\_r_j^{-1}$, where $v$ is some vocabulary word. A {\em context configuration} is then simply a set of individual context bags, e.g., $R=\{r_i,r_j,r_k\}$, also labelled as $R$: $r_i+r_j+r_k$. We call a configuration consisting of $K$ individual context bags a $K$-set configuration (e.g., in this example, $R$ is a $3$-set configuration).\footnote{A note on the nomenclature and notation: Each context configuration may be seen as a set of context bags, as it does not allow for repetition of its constituent context bags. For simplicity and clarity of presentation, we use dependency relation types (e.g., $r_i$ = \texttt{amod}, $r_j$ = \texttt{acl}) as labels for context bags. The reader has to be aware that a configuration $R=\{r_i,r_j,r_k\}$ is not by any means a set of relation types/names, but is in fact a multiset of all $(word, context)$ pairs belonging to the corresponding context bags labelled with $r_i$, $r_j$, $r_k$.}
\begin{figure}[t]
\centering
\includegraphics[width=1.0\linewidth]{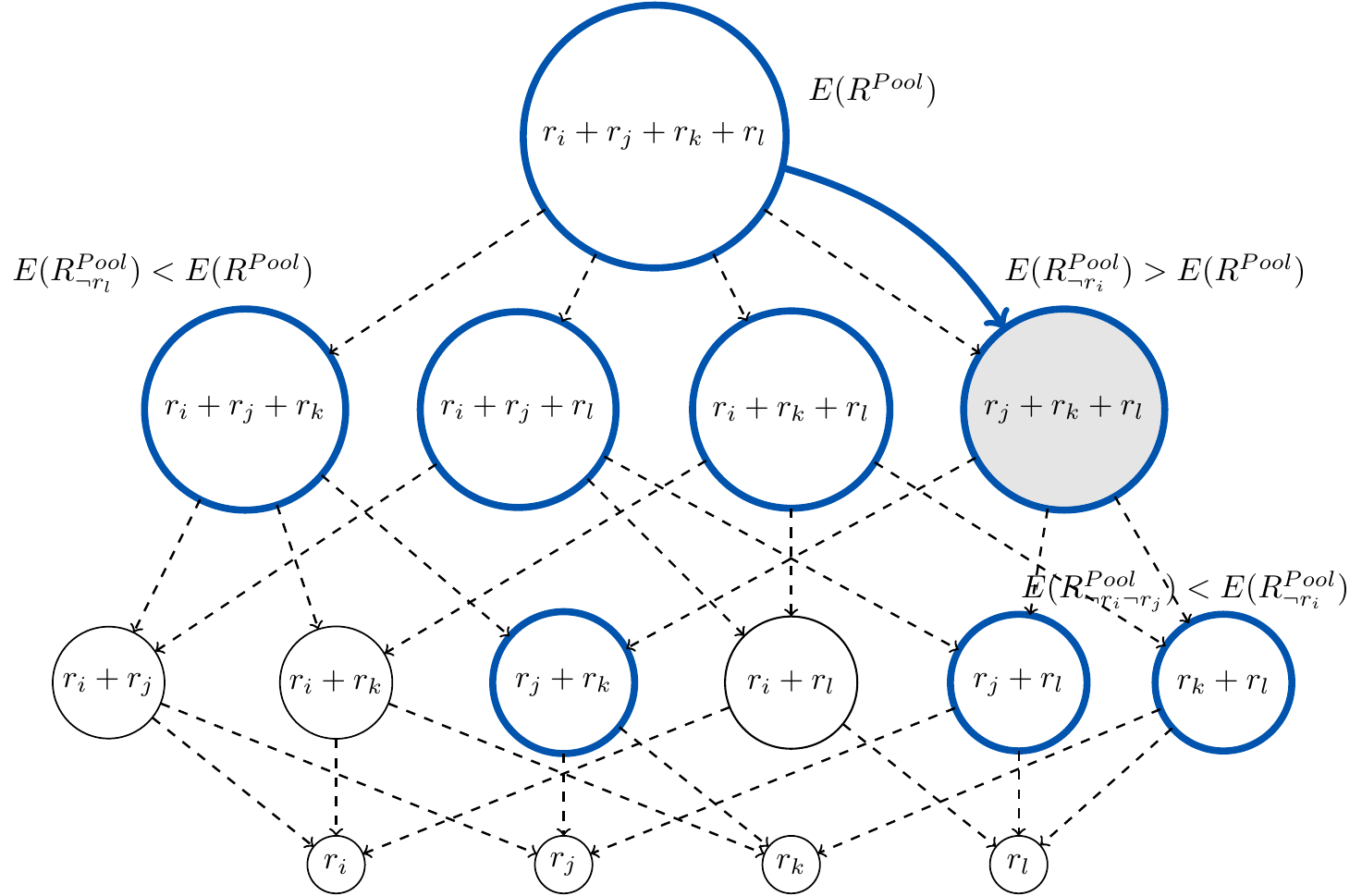}
\vspace{-0.7em}
\caption{An illustration of Alg.~1. The search space is presented as a DAG with direct links between origin configurations (e.g., $r_i+r_j+r_k$) and all its children configurations obtained by removing exactly one individual bag from the origin (e.g., $r_i+r_j$, $r_j+r_k$). After automatically constructing the initial pool (line 1), the entry point of the algorithm is the $R^{Pool}$ configuration (line 2). Thicker blue circles denote visited configurations, while the gray circle denotes the best configuration found.}
\vspace{-0.7em}
\label{fig:alggraph}
\end{figure}
\begin{figure}
\removelatexerror
\begin{footnotesize}
\begin{algorithm}[H]
 \SetKwData{Left}{left}\SetKwData{This}{this}\SetKwData{Up}{up}
 \SetKwFunction{Union}{Union}\SetKwFunction{FindCompress}{FindCompress}
 \SetKwInOut{Input}{Input}\SetKwInOut{Output}{Output}
 \Input{Set of $M$ individual context bags: $S=\{r'_1,r'_2,\ldots,r'_M\}$}
 {\bf build}: {\em pool} of  those $K\leq M$ candidate individual context bags $\{r_1,\ldots,r_K\}$ for which $E(r_i) >= threshold, i \in \{1, \ldots, M\}$, where $E(\cdot)$ is a fitness function.\\
 {\bf build:} $K$-set configuration $R^{Pool} = \{r_1,\ldots,r_K\}$ \;
 {\bf initialize:} (1) set of candidate configurations $\mathbf{R} = \{R^{Pool}\}$ ; (2) current level $l=K$ ; (3) best configuration $R_o =\emptyset$ \;
 {\bf search}: \\
 \Repeat{$l==0$ or $\mathbf{R}==\emptyset$}
 { 
 $\mathbf{R}_n \leftarrow \emptyset$ \;
 $R_o \leftarrow \underset{R \in \mathbf{R} \cup \{R_o\}} {\operatorname{arg\,max}} \hspace{0.3em}E(R)$ \;
 \ForEach{$R \in \mathbf{R}$}{
		\ForEach{$r_i \in R$}{
		{\bf build} new $(l-1)$-set context configuration $R_{\neg r_i}=R-\{r_i\}$ \;
		\If{$E(R_{\neg r_i})\geq E(R)$}
		{
			$\mathbf{R_n} \leftarrow \mathbf{R_n} \cup \{R_{\neg r_i}\}$ \;		
		}
	}
 }
 $l \leftarrow l-1$ \;
 $\mathbf{R} \leftarrow \mathbf{R_n}$ \;
 }
 \Output{Best configuration $R_o$}
 \caption{\small Best Configuration Search}
\end{algorithm}
\end{footnotesize}
\vspace{-0.6em}
\end{figure}

Although a brute-force exhaustive search over all possible configurations is possible in theory and for small pools (e.g., for adjectives, see Tab.~\ref{tab:pool}), it becomes challenging or practically infeasible for large pools and large training data. For instance, based on the pool from Tab.~\ref{tab:pool}, the search for the optimal configuration would involve trying out $2^{10}-1=1023$ configurations for nouns 
(i.e., training 1023 different word representation models). Therefore, to reduce the number of visited configurations, we present a simple heuristic search algorithm inspired by beam search \cite{Pearl:1984}.


\subsection{Class-Specific Configuration Search}
\label{ss:pools}

Alg.~1 provides a high-level overview of the algorithm. An example of its flow is given in Fig.~\ref{fig:alggraph}. 
Starting from $S$, the set of all possible $M$ individual context bags, the algorithm automatically detects the subset $S_K \subseteq S$, $|S_K|=K$, of candidate individual bags that are used as the initial pool (line 1 of Alg.~1). The selection is based on some fitness (goal) function $E$. In our setup, $E(R)$ is Spearman's $\rho$ correlation with human judgment scores obtained on the development set after training the word representation model with the configuration $R$. The selection step relies on a simple threshold: we use a threshold of $\rho\geq 0.2$ without any fine-tuning in all experiments with all word classes.

We find this step to facilitate efficiency at a minor cost for accuracy.  For example, since \texttt{amod} denotes an adjectival modifier of a noun, an efficient search procedure may safely remove this bag from the pool of candidate bags for verbs.

The search algorithm then starts from the full $K$-set $R^{Pool}$ configuration (line 3) and tests $K$ $(K-1)$-set configurations where exactly one individual bag $r_i$ is removed to generate each such configuration (line 10). It then retains only the set of configurations that score higher than the origin $K$-set configuration (lines 11-12, see Fig.~\ref{fig:alggraph}). Using this principle, it continues searching only over lower-level $(l-1)$-set configurations that further improve performance over their $l$-set origin configuration. It stops if it reaches the lowest level or if it cannot improve the goal function any more (line 15). The best scoring configuration is returned (n.b., not guaranteed to be the global optimum).

In our experiments with this heuristic, the search for the optimal configuration for verbs is performed only over 13 $1$-set configurations plus 26 other configurations (39 out of 133 possible configurations).\footnote{The total is 133 as we have to include 6 additional $1$-set configurations that have to be tested (line 1 of Alg.~1) but are not included in the initial pool for verbs (line 2).}  For nouns, the advantage of the heuristic is even more dramatic: only 104 out of 1026 possible configurations were considered during the search.\footnote{We also experimented with a less conservative variant which does not stop when lower-level configurations do not improve $E$; it instead follows the path of the best-scoring lower-level configuration even if its score is lower than that of its origin. As we do not observe any significant improvement with this variant, we opt for the faster and simpler one.}

\section{Experimental Setup}
\label{s:exp}
\subsection{Implementation Details}

\paragraph{Word Representation Model} 
We experiment with SGNS \cite{Mikolov:2013nips}, the standard and very robust choice in vector space modeling \cite{Levy:2015tacl}. In all experiments we use \texttt{word2vecf}, a reimplementation of \texttt{word2vec} able to learn from arbitrary $(word, context)$ pairs.\footnote{https://bitbucket.org/yoavgo/word2vecf} For details concerning the implementation, we refer the reader to \cite{Goldberg:2014arxiv,Levy:2014acl}.

 The SGNS preprocessing scheme was replicated from \cite{Levy:2014acl,Levy:2015tacl}. After lowercasing, all words and contexts that appeared less than 100 times were filtered. When considering all dependency types, the vocabulary spans approximately 185K word types.\footnote{SGNS for all models was trained using stochastic gradient descent and standard settings: $15$ negative samples, global learning rate: $0.025$, subsampling rate: $1e-4$, $15$ epochs.} Further, all representations were trained with $d=300$ (very similar trends are observed with $d=100,500$).
 
The same setup was used in prior work \cite{Schwartz:2016naacl,Vulic:2016acluniversal}. Keeping the representation model fixed across experiments and varying only the context type allows us to attribute any differences in results to a sole factor: the context type. We plan to experiment with other representation models in future work.

\paragraph{Universal Dependencies as Labels} 
The adopted UD scheme leans on the universal Stanford dependencies \cite{Marneffe:2014lrec} complemented with the universal POS tagset \cite{Petrov:2012lrec}. It is straightforward to ``translate'' previous annotation schemes to UD \cite{Marneffe:2014lrec}. Providing a consistently annotated inventory of categories for similar syntactic constructions across languages, the UD scheme facilitates representation learning in languages other than English, as shown in  \cite{Vulic:2016acluniversal,Vulic:2017eacl}.

\paragraph{Individual Context Bags} 
\vspace{-0.0em}
Standard post-parsing steps are performed in order to obtain an initial list of individual context bags for our algorithm: (1) Prepositional arcs are collapsed  (\cite{Levy:2014acl,Vulic:2016acluniversal}, see Fig.~\ref{fig:rex}). Following this procedure, all pairs where the relation $r$ has the form \texttt{prep:X} (where \texttt{X} is a preposition) are subsumed to a context bag labelled \texttt{prep}; (2) Similar labels are merged into a single label (e.g., direct (\texttt{dobj}) and indirect objects (\texttt{iobj}) are merged into \texttt{obj}); (3) Pairs with infrequent and uninformative labels are removed (e.g., \texttt{punct}, \texttt{goeswith}, \texttt{cc}).

Coordination-based contexts are extracted as in prior work \cite{Schwartz:2016naacl}, distinguishing between left and right contexts extracted from the \texttt{conj} relation; the label for this bag is \texttt{conjlr}. We also utilise the variant that does not make the distinction, labeled \texttt{conjll}. If both are used, the label is simply \texttt{conj=conjlr+conjll}.\footnote{Given the coordination structure {\em boys and girls}, \texttt{conjlr} training pairs are {\em (boys, girls\_conj), (girls, $boys\_conj^{-1})$}, while \texttt{conjll} pairs are {\em (boys, girls\_conj), (girls, $boys\_conj)$}.}

Consequently, the individual context bags we use in all experiments are: \texttt{subj}, \texttt{obj}, \texttt{comp}, \texttt{nummod}, \texttt{appos}, \texttt{nmod}, \texttt{acl}, \texttt{amod}, \texttt{prep}, \texttt{adv}, \texttt{compound}, \texttt{conjlr}, \texttt{conjll}.

\subsection{Training and Evaluation}

We run the algorithm for context configuration selection only once, with the SGNS training setup described below. Our main evaluation setup is presented below, but the learned configurations are tested in additional setups, detailed in Sect.~\ref{s:results}.

\paragraph{Training Data} Our training corpus is the cleaned and tokenised English Polyglot Wikipedia data \cite{AlRfou:2013conll},\footnote{https://sites.google.com/site/rmyeid/projects/polyglot} consisting of approximately 75M sentences and 1.7B word tokens. 
The Wikipedia data were POS-tagged with universal POS (UPOS) tags \cite{Petrov:2012lrec} using the state-of-the art TurboTagger \cite{Martins:2013acl}.\footnote{http://www.cs.cmu.edu/\textasciitilde ark/TurboParser/} The parser was trained using default settings (SVM MIRA with 20 iterations, no further parameter tuning) on the {\sc train+dev} portion of the UD treebank annotated with UPOS tags. The data were then parsed with UD using the graph-based Mate parser v3.61 \cite{Bohnet:2010coling}\footnote{https://code.google.com/archive/p/mate-tools/} with standard settings on {\sc train+dev} of the UD treebank.


\paragraph{Evaluation} 
We experiment with the verb pair (222 pairs), adjective pair (111 pairs), and noun pair (666 pairs) portions of SimLex-999. We report Spearman's $\rho$ correlation between the ranks
derived from the scores of the evaluated models and the human scores. Our evaluation setup is borrowed from \newcite{Levy:2015tacl}: we perform 2-fold cross-validation, where the context configurations are optimised on a development set, separate from the unseen test data. Unless stated otherwise, the reported scores are always the averages of the 2 runs, computed in the standard fashion by applying the cosine similarity to the vectors of words participating in a pair.

\subsection{Baselines}

\paragraph{Baseline Context Types} We compare the context configurations found by Alg.~1 against baseline contexts from prior work:\\ 
{\bf -  BOW}: Standard bag-of-words contexts. \\
{\bf -  POSIT}: Positional contexts \cite{Schutze:1993acl,Levy:2014conll,Ling:2015naacl}, which enrich BOW with information on the sequential position of each context word. Given the example from Fig.~\ref{fig:rex}, POSIT with the window size $2$ extracts the following contexts for {\em discovers}: {\em Australian\_-2}, {\em scientist\_-1}, {\em stars\_+2}, {\em with\_+1}. \\
{\bf -  DEPS-All}: All dependency links without any context selection, extracted from dependency-parsed data with prepositional arc collapsing. \\
{\bf -  COORD}: Coordination-based contexts are used as fast lightweight contexts for improved representations of adjectives and verbs \cite{Schwartz:2016naacl}. This is in fact the \texttt{conjlr} context bag, a subset of DEPS-All. \\
{\bf -  SP}: Contexts based on symmetric patterns (SPs, \cite{Davidov:2006acl,Schwartz:2015conll}). For example, if the word X and  the word Y appear in the lexico-syntactic symmetric pattern ``X or Y'' in the SGNS training corpus, then Y is an SP context instance for X, and vice versa. 

The development set was used to tune the window size for BOW and POSIT (to 2) and the parameters of the SP extraction algorithm.\footnote{The SP extraction algorithm is available online: \\ homes.cs.washington.edu/$\sim$roysch/software/dr06/dr06.html}

\paragraph{Baseline Greedy Search Algorithm} We also compare our search algorithm to its greedy variant: at each iteration of lines 8-12 in Alg.~1, $R_n$ now keeps only the best configuration of size $l-1$ that perform better than the initial configuration of size $l$, instead of all such configurations.

\begin{table}
\centering
\begin{footnotesize}
\def\arraystretch{1.0}
\begin{tabularx}{1.0\linewidth}{l XXX}
\toprule
{\bf Context Group} & {Adj} & {Verb} & {Noun} \\
\cmidrule(lr){1-1} \cmidrule(lr){2-4}
{\texttt{conjlr} (A+N+V)} & \cellcolor{Gray}{0.415} & \cellcolor{Gray}{0.281} & \cellcolor{Gray}{0.401} \\
{\texttt{obj} (N+V)} & {-0.028} & \cellcolor{Gray}{0.309} & \cellcolor{Gray}{0.390} \\
{\texttt{prep} (N+V)} & {0.188} & \cellcolor{Gray}{0.344} & \cellcolor{Gray}{0.387} \\
{\texttt{amod} (A+N)} & \cellcolor{Gray}{0.479} & {0.058} & \cellcolor{Gray}{0.398} \\
{\texttt{compound} (N)} & {-0.124} & {-0.019} & \cellcolor{Gray}{0.416} \\
{\texttt{adv} (V)} & {0.197} & \cellcolor{Gray}{0.342} & {0.104} \\
{\texttt{nummod} (-)} & {-0.142} & {-0.065} & {0.029} \\
\bottomrule
\end{tabularx}
\vspace{-0.3em}
\caption{2-fold cross-validation results for an illustrative selection of individual context bags. Results are presented for the noun, verb and adjective subsets of SimLex-999. Values in parentheses denote the class-specific initial pools to which each context is selected based on its $\rho$ score (line 1 of Alg.~1).}
\label{tab:single}
\end{footnotesize}
\vspace{1.5em}
\centering
\begin{footnotesize}
\def\arraystretch{0.75}
\begin{tabularx}{1.0\linewidth}{smb}
\toprule
{Adjectives} & {Verbs} & {Nouns} \\
\midrule
\texttt{amod, conjlr, conjll} & \texttt{prep, acl, obj, comp, adv, conjlr, conjll} & \texttt{amod, prep, compound, subj, obj, appos, acl, nmod, conjlr, conjll} \\
\bottomrule
\end{tabularx}
\vspace{-0.7em}
\caption{Automatically constructed initial pools of candidate bags for each word class (Sect.~\ref{ss:pools}).}
\label{tab:pool}
\end{footnotesize}
\vspace{-0.6em}
\end{table}
\begin{table*}
\centering
\begin{tabular}{cc}
\begin{footnotesize}
\def\arraystretch{0.9}
\begin{tabularx}{0.42\linewidth}{l X}
\toprule
{\bf Baselines} & {\bf (Verbs)} \\
\cmidrule(lr){2-2}
\cmidrule(lr){1-1}
{BOW (\texttt{win}=2)} & {0.336}  \\
{POSIT (\texttt{win}=2)} & {0.345}  \\
{COORD (\texttt{conjlr})} & {0.283}  \\
{SP} & {0.349} \\
{DEPS-All} & {0.344}  \\
\midrule
{\bf Configurations: Verbs} & {}  \\
\cmidrule(lr){1-1}
{\scriptsize \texttt{POOL-ALL}} & \cellcolor{Gray}{0.379}  \\
{\scriptsize \texttt{prep+acl+obj+adv+conj}} & \cellcolor{Gray}{0.393}  \\
{\scriptsize \texttt{prep+acl+obj+comp+conj}} & {0.344}  \\
{\scriptsize \texttt{prep+obj+comp+adv+conj}} & \cellcolor{Gray}{\em 0.391}$^\dagger$  \\
{\scriptsize \texttt{prep+acl+adv+conj}} (BEST) & \cellcolor{Gray}{\bf 0.409}  \\
{\scriptsize \texttt{prep+acl+obj+adv}} & \cellcolor{Gray}{0.392}   \\
{\scriptsize \texttt{prep+acl+adv}} & \cellcolor{Gray}{0.407}  \\
{\scriptsize \texttt{prep+acl+conj}} & \cellcolor{Gray}{0.390}  \\
{\scriptsize \texttt{acl+obj+adv+conj}} & {0.345} \\
{\scriptsize \texttt{acl+obj+adv}} & \cellcolor{Gray}{0.385} \\
\bottomrule
\end{tabularx}
\vspace{-0.0em}
\end{footnotesize}
\vspace{-0.0em}
&
\centering
\begin{footnotesize}
\def\arraystretch{0.9}
\begin{tabularx}{0.52\linewidth}{l X}
\toprule
{\bf Baselines} & {{\bf (Nouns)}} \\
\cmidrule(lr){2-2}
\cmidrule(lr){1-1}
{BOW (\texttt{win}=2)} & {0.435}  \\
{POSIT (\texttt{win}=2)} & {0.437}  \\
{COORD (\texttt{conjlr})} & {0.392}  \\
{SP} & {0.372} \\
{DEPS-All} & {0.441}  \\
\midrule
{\bf Configurations: Nouns} & {}  \\
\cmidrule(lr){1-1}
{\scriptsize \texttt{POOL-ALL}} & \cellcolor{Gray}{0.469}  \\
{\scriptsize \texttt{amod+subj+obj+appos+compound+nmod+conj}} & \cellcolor{Gray}{0.478}  \\
{\scriptsize \texttt{amod+subj+obj+appos+compound+conj}} & \cellcolor{Gray}{0.487}  \\
{\scriptsize \texttt{amod+subj+obj+appos+compound+conjlr}} & \cellcolor{Gray}{\em 0.476}$^\dagger$  \\
{\scriptsize \texttt{amod+subj+obj+compound+conj}} (BEST) & \cellcolor{Gray}{\bf 0.491}  \\
{\scriptsize \texttt{amod+subj+obj+appos+conj}} & \cellcolor{Gray}{0.470}  \\
{\scriptsize \texttt{subj+obj+compound+conj}} & \cellcolor{Gray}{0.479}  \\
{\scriptsize \texttt{amod+subj+compound+conj}} & \cellcolor{Gray}{0.481}  \\
{\scriptsize \texttt{amod+subj+obj+compound}} & \cellcolor{Gray}{0.478}  \\
{\scriptsize \texttt{amod+obj+compound+conj}} & \cellcolor{Gray}{0.481}  \\
\bottomrule
\end{tabularx}
\end{footnotesize}
\vspace{-0.0em}
\end{tabular}
\caption{Results on the SimLex-999 test data over (a) {\bf verbs} and (b) {\bf nouns} subsets. Only a selection of context configurations optimised for verb and noun similarity are shown. POOL-ALL denotes a configuration where all individual context bags from the verbs/nouns-oriented pools (see Table~\ref{tab:pool}) are used. BEST denotes the best performing configuration found by Alg.~1. Other configurations visited by Alg.~1 that score higher than the best scoring baseline context type for each word class are in gray. Scores obtained using a greedy search algorithm instead of Alg.~1 are in italic, marked with a cross ($\dagger$).}
\vspace{-0.5em}
\label{tab:verbsnouns}
\end{table*}

\section{Results and Discussion}
\label{s:results}
\subsection{Main Evaluation Setup}
\label{ss:main-eval}

\paragraph{Not All Context Bags are Created Equal}
First, we test the performance of {\em individual} context bags across SimLex-999 adjective, verb, and noun subsets. Besides providing insight on the intuition behind context selection, these findings are important for the automatic selection of class-specific pools (line 1 of Alg.~1). The results are shown in Tab.~\ref{tab:single}.

The experiment supports our intuition (see Sect.~\ref{ss:pools}): some context bags are definitely not useful for some classes and may be safely removed when performing the class-specific SGNS training. For instance, the \texttt{amod} bag is indeed important for adjective and noun similarity, and at the same time it does not encode any useful information regarding verb similarity. \texttt{compound} is, as expected, useful only for nouns. Tab.~\ref{tab:single} also suggests that some context bags (e.g., \texttt{nummod}) do not encode any informative contextual evidence regarding similarity, therefore they can be discarded. The initial results with individual context bags help to reduce the pool of candidate bags (line 1 in Alg.~1), see Tab.~\ref{tab:pool}.
\begin{table}[t]
\centering
\begin{footnotesize}
\def\arraystretch{0.9}
\begin{tabularx}{1.0\linewidth}{l X}
\toprule
{\bf Baselines} & {\bf (Adjectives)} \\
\cmidrule(lr){2-2}
\cmidrule(lr){1-1}
{BOW (\texttt{win}=2)} & {0.489}  \\
{POSIT (\texttt{win}=2)} & {0.460}  \\
{COORD (\texttt{conjlr})} & {0.407}  \\
{SP} & {0.395} \\
{DEPS-All} & {0.360}  \\
\midrule
{\bf Configurations: Adjectives} & {}  \\
\cmidrule(lr){1-1}
{\scriptsize \texttt{POOL-ALL: amod+conj}} (BEST) & \cellcolor{Gray}{\bf 0.546}$^\dagger$  \\
{\scriptsize \texttt{amod+conjlr}} & \cellcolor{Gray}{0.527}  \\
{\scriptsize \texttt{amod+conjll}} & \cellcolor{Gray}{0.531}  \\
{\scriptsize \texttt{conj}} & {0.470}  \\
\bottomrule
\end{tabularx}
\vspace{-0.5em}
\caption{Results on the SimLex-999 {\bf adjectives} subset with adjective-specific configurations.}
\label{tab:adj}
\end{footnotesize}
\vspace{-0.3em}
\end{table}


\paragraph{Searching for Improved Configurations}
Next, we test if we can improve class-specific representations by selecting class-specific configurations. 
Results are summarised in Tables~\ref{tab:verbsnouns} and \ref{tab:adj}. Indeed, class-specific configurations yield better representations, as is evident from the scores: the improvements with the best class-specific configurations found by Alg.~1 are approximately 6 $\rho$ points for adjectives, 6 points for verbs, and 5 points for nouns over the best baseline for each class. 

The improvements are visible even with configurations that simply pool all candidate individual bags (POOL-ALL), without running Alg.~1 beyond line 1. However, further careful context selection, i.e., traversing the configuration space using Alg.~1 leads to additional improvements for V and N (gains of 3 and 2.2 $\rho$ points). 
Very similar improved scores are achieved with a variety of configurations (see Tab.~\ref{tab:verbsnouns}), especially in the neighbourhood of the best configuration found by Alg.~1. This indicates that the method is quite robust: even sub-optimal\footnote{The term \textit{optimal} here and later in the text refers to the best configuration returned by our algorithm.} solutions result in improved class-specific representations. Furthermore, our algorithm is able to find better configurations for verbs and nouns compared to its greedy variant. Finally, our algorithm generalises well: the best scoring configuration on the dev set is always the best one on the test set.

\paragraph{Training: Fast and/or Accurate?}
Carefully selected configurations are also likely to reduce SGNS training times. Indeed, the configuration-based model trains on only 14\% (A), 26.2\% (V), and 33.6\% (N) of all dependency-based contexts. The training times and statistics for each context type are displayed in Tab.~\ref{tab:speed}. All models were trained using parallel training on 10 Intel(R) Xeon(R) E5-2667 2.90GHz processors. The results indicate that class-specific configurations are not as lightweight and fast as SP or COORD contexts \cite{Schwartz:2016naacl}. However, they also suggest that such configurations provide a good balance between accuracy and speed: they reach peak performances for each class, outscoring all baseline context types (including SP and COORD), while training is still much faster than with ``heavyweight'' context types such as BOW, POSIT or DEPS-All.

Now that we verified the decrease in training time our algorithm provides for the final training, it makes sense to ask whether the configurations it finds are valuable \textit{in other setups}. This will make the fast training of practical importance.

\begin{table}[t]
\centering
\begin{footnotesize}
\def\arraystretch{1.0}
\begin{tabularx}{1.0\linewidth}{l XX}
\toprule
{\bf Context Type} & {Training Time} & {\# Pairs} \\
\cmidrule(lr){1-1} \cmidrule(lr){2-2} \cmidrule(lr){3-3}
{BOW (\texttt{win}=2)} & {179mins 27s} & {5.974G}\\
{POSIT (\texttt{win}=2)} & {190mins 12s} & {5.974G}\\
{COORD (\texttt{conjlr})} & {4mins 11s} & {129.69M} \\
{SP} & {1mins 29s} & {46.37M} \\
{DEPS-All} & {103mins 35s} & {3.165G} \\
\midrule
{BEST-ADJ} & {14mins 5s} & {447.4M} \\
{BEST-VERBS} & {29mins 48s} & {828.55M} \\
{BEST-NOUNS} & {41mins 14s} & {1.063G} \\
\bottomrule
\end{tabularx}
\vspace{-0.3em}
\caption{Training time (wall-clock time reported) in minutes for SGNS ($d=300$) with different context types. BEST-* denotes the best scoring configuration for each class found by Alg.~1. {\em \#Pairs} shows a total number of pairs used in SGNS training for each context type.}
\label{tab:speed}
\end{footnotesize}
\vspace{-0.2em}
\end{table}


\subsection{Generalisation: Configuration Transfer}

\paragraph{Another Training Setup} We first test whether the context configurations learned in Sect.~\ref{ss:main-eval} are useful when SGNS is trained in another English setup \cite{Schwartz:2016naacl}, with more training data and other annotation and parser choices, while evaluation is still performed on SimLex-999.

\begin{table}[t]
\centering
\begin{footnotesize}
\def\arraystretch{1.0}
\begin{tabularx}{1.0\linewidth}{l XXXX}
\toprule
{\bf Context Type} & {Adj} & {Verbs} & {Nouns} & {All}\\
\cmidrule(lr){1-1} \cmidrule(lr){2-2} \cmidrule(lr){3-3} \cmidrule(lr){4-4} \cmidrule(lr){5-5}
{BOW (\texttt{win}=2)} & {0.604} & {0.307} & {0.501} & {0.464}\\
{POSIT (\texttt{win}=2)} & {0.585} & {0.400} & {0.471} & {0.469}\\
{COORD (\texttt{conjlr})} & {0.629} & {0.413} & {0.428} & {0.430} \\
{SP} & {0.649} & {\bf 0.458} & {0.414} & {0.444}\\
{DEPS-All} & {0.574} & {0.389} & {0.492} & {0.464}\\
\midrule
{BEST-ADJ} & {\bf 0.671} & {0.348} & {0.504} & {0.449}\\
{BEST-VERBS} & {0.392} & {0.455} & {0.478} & {0.448}\\
{BEST-NOUNS} & {0.581} & {0.327} & {\bf 0.535} & {0.489}\\
\midrule
{BEST-ALL} & {0.616} & {0.402} & {0.519} & {\bf 0.506} \\
\bottomrule
\end{tabularx}
\vspace{-0.2em}
\caption{Results on the A/V/N SimLex-999 subsets, and on the entire set ({\em All}) in the setup from \protect\newcite{Schwartz:2016naacl}. $d=500$. BEST-* are again the best class-specific configs returned by Alg.~1.}
\label{tab:roy}
\end{footnotesize}
\vspace{-0.3em}
\end{table}

In this setup the training corpus is the 8B words corpus generated by the \texttt{word2vec} script.\footnote{code.google.com/p/word2vec/source/browse/trunk/}  A preprocessing step now merges common word pairs and triplets to expression tokens (e.g., {\em Bilbo\_Baggins}). The corpus is parsed with labelled Stanford dependencies \cite{Marneffe:2008sd} using the Stanford POS Tagger \cite{Toutanova:2003naacl} and the stack version of the MALT parser \cite{Goldberg:2012coling}. SGNS preprocessing and parameters are also replicated; we now train $500$-dim embeddings as in prior work.\footnote{The ``translation'' from labelled Stanford dependencies into UD is performed using the mapping from \newcite{Marneffe:2014lrec}, e.g., \texttt{nn} is mapped into \texttt{compound}, and \texttt{rcmod}, \texttt{partmod}, \texttt{infmod} are all mapped into one bag: \texttt{acl}.}

Results are presented in Tab.~\ref{tab:roy}. The imported class-specific configurations, computed using a much smaller corpus (Sect.~\ref{ss:main-eval}), again outperform competitive baseline context types for adjectives and nouns. The BEST-VERBS configuration is outscored by SP, but the margin is negligible. We also evaluate another configuration found using Alg.~1 in Sect.~\ref{ss:main-eval}, which targets the overall improved performance without any finer-grained division to classes (BEST-ALL). This configuration ({\small \texttt{amod+subj+obj+compound+prep+adv+conj}}) outperforms all baseline models on the entire benchmark. 
Interestingly, the non-specific BEST-ALL configuration falls short of A/V/N-specific configurations for each class. This unambiguously implies that the ``trade-off'' configuration targeting all three classes at the same time differs from specialised class-specific configurations.

\begin{table}[t]
\centering
\begin{footnotesize}
\def\arraystretch{1.0}
\begin{tabularx}{1.0\linewidth}{l XXX}
\toprule
{\bf Context Type} & {Adj-Q} & {Verb-Q} & {Noun-Q} \\
\cmidrule(lr){1-1} \cmidrule(lr){2-2} \cmidrule(lr){3-3} \cmidrule(lr){4-4} 
{BOW (\texttt{win}=2)} & {31/41} & {14/19} & {16/19} \\
{POSIT (\texttt{win}=2)} & {\bf 32/41} & {13/19} & {15/19} \\
{COORD (\texttt{conjlr})} & {26/41} & {11/19} & {\hspace{0.5em}8/19} \\
{SP} & {26/41} & {11/19} & {12/19} \\
{DEPS-All} & {31/41} & {14/19} & {16/19} \\
\midrule
{BEST-ADJ} & {\bf 32/41} & {12/19} & {15/19} \\
{BEST-VERBS} & {24/41} & {\bf 15/19} & {16/19} \\
{BEST-NOUNS} & {30/41} & {14/19} & {\bf 17/19} \\
\bottomrule
\end{tabularx}
\vspace{-0.2em}
\caption{Results on the A/V/N TOEFL question subsets. The reported scores are in the following form: {\em correct\_answers/overall\_questions}. \textit{Adj-Q} refers to the subset of TOEFL questions targeting adjectives; similar for \textit{Verb-Q} and \textit{Noun-Q}. BEST-* refer to the best class-specific configurations from Tab.~3 and Tab.~4.}
\label{tab:toefl}
\end{footnotesize}
\vspace{-0.3em}
\end{table}

\paragraph{Experiments on Other Languages}
We next test whether the optimal context configurations computed in Sect.~\ref{ss:main-eval} with English training data are also useful for other languages. For this, we train SGNS models on the Italian (IT) and German (DE) Polyglot Wikipedia corpora with those configurations, and evaluate on the IT and DE multilingual SimLex-999 \cite{Leviant:2015arxiv}.\footnote{http://leviants.com/ira.leviant/MultilingualVSMdata.html} 

Our results demonstrate similar patterns as for English, and indicate that our framework can be easily applied to other languages. For instance, the BEST-ADJ configuration (the same configuration as in Tab.~4 and Tab.~7) yields an improvement of 8 $\rho$ points and 4 $\rho$ points over the strongest adjectives baseline in IT and DE, respectively. We get similar improvements for nouns (IT: 3 $\rho$ points, DE: 2 $\rho$ points), and verbs (IT: 2, DE: 4).

\paragraph{TOEFL Evaluation} We also verify that the selection of class-specific configurations (Sect.~\ref{ss:main-eval}) is useful beyond the core SimLex evaluation. For this aim, we evaluate on the A, V, and N TOEFL questions \cite{Landauer:1997pr}. The results are summarised in Tab.~\ref{tab:toefl}. Despite the limited size of the TOEFL dataset, we observe positive trends in the reported results (e.g., V-specific configurations yield a small gain on verb questions), showcasing the potential of class-specific training in this task.



\vspace{-0.0em}
\section{Conclusion and Future Work}
\label{s:conclusion}
We have presented a novel framework for selecting class-specific context configurations which yield improved representations for prominent word classes: adjectives, verbs, and nouns. Its design and dependence on the Universal Dependencies annotation scheme makes it applicable in different languages. We have proposed an algorithm that is able to find a suitable class-specific configuration while making the search over the large space of possible context configurations computationally feasible. Each word class requires a different class-specific configuration to produce improved results on the class-specific subset of SimLex-999 in English, Italian, and German. We also show that the selection of context configurations is robust as once learned configuration may be effectively transferred to other data setups, tasks, and languages without additional retraining or fine-tuning.

In future work, we plan to test the framework with finer-grained contexts, investigating beyond POS-based word classes and dependency links. Exploring more sophisticated algorithms that can efficiently search richer configuration spaces is also an intriguing direction. Another research avenue is application of the context selection idea to other representation models beyond SGNS tested in this work, and experimenting with assigning weights to context subsets. Finally, we plan to test the portability of our approach to more languages.

\section*{Acknowledgments}
This work is supported by the ERC Consolidator Grant LEXICAL: Lexical Acquisition Across Languages (no 648909). Roy Schwartz was supported by the Intel Collaborative Research Institute for Computational Intelligence (ICRI-CI). The authors are grateful to the anonymous reviewers for their helpful and constructive suggestions.

\bibliography{acl2017_refs}

\begin{thebibliography}{}
\expandafter\ifx\csname natexlab\endcsname\relax\def\natexlab#1{#1}\fi

\bibitem[{Al-Rfou et~al.(2013)Al-Rfou, Perozzi, and Skiena}]{AlRfou:2013conll}
Rami Al-Rfou, Bryan Perozzi, and Steven Skiena. 2013.
\newblock \href{http://www.aclweb.org/anthology/W13-3520}{Polyglot:
  {D}istributed word representations for multilingual {NLP}}.
\newblock In {\em CoNLL\/}. pages 183--192.
\newblock
  \href{http://www.aclweb.org/anthology/W13-3520}{http://www.aclweb.org/anthology/W13-3520}.

\bibitem[{Bansal et~al.(2014)Bansal, Gimpel, and Livescu}]{Bansal:2014acl}
Mohit Bansal, Kevin Gimpel, and Karen Livescu. 2014.
\newblock \href{http://www.aclweb.org/anthology/P14-2131}{Tailoring continuous
  word representations for dependency parsing}.
\newblock In {\em ACL\/}. pages 809--815.
\newblock
  \href{http://www.aclweb.org/anthology/P14-2131}{http://www.aclweb.org/anthology/P14-2131}.

\bibitem[{Baroni et~al.(2014)Baroni, Dinu, and Kruszewski}]{Baroni:2014acl}
Marco Baroni, Georgiana Dinu, and Germ{\'{a}}n Kruszewski. 2014.
\newblock \href{http://www.aclweb.org/anthology/P14-1023}{Don't count, predict!
  {A} systematic comparison of context-counting vs. context-predicting semantic
  vectors}.
\newblock In {\em ACL\/}. pages 238--247.
\newblock
  \href{http://www.aclweb.org/anthology/P14-1023}{http://www.aclweb.org/anthology/P14-1023}.

\bibitem[{Baroni et~al.(2010)Baroni, Murphy, Barbu, and
  Poesio}]{Baroni:2010cogsci}
Marco Baroni, Brian Murphy, Eduard Barbu, and Massimo Poesio. 2010.
\newblock \href{https://doi.org/10.1111/j.1551-6709.2009.01068.x}{{Strudel: A}
  corpus-based semantic model based on properties and types}.
\newblock {\em Cognitive Science\/} pages 222--254.
\newblock
  \href{https://doi.org/10.1111/j.1551-6709.2009.01068.x}{https://doi.org/10.1111/j.1551-6709.2009.01068.x}.

\bibitem[{Bohnet(2010)}]{Bohnet:2010coling}
Bernd Bohnet. 2010.
\newblock \href{http://www.aclweb.org/anthology/C10-1011}{Top accuracy and fast
  dependency parsing is not a contradiction}.
\newblock In {\em COLING\/}. pages 89--97.
\newblock
  \href{http://www.aclweb.org/anthology/C10-1011}{http://www.aclweb.org/anthology/C10-1011}.

\bibitem[{Chen and Manning(2014)}]{Chen:2014emnlp}
Danqi Chen and Christopher~D. Manning. 2014.
\newblock \href{http://www.aclweb.org/anthology/D14-1082}{A fast and accurate
  dependency parser using neural networks}.
\newblock In {\em EMNLP\/}. pages 740--750.
\newblock
  \href{http://www.aclweb.org/anthology/D14-1082}{http://www.aclweb.org/anthology/D14-1082}.

\bibitem[{Collobert et~al.(2011)Collobert, Weston, Bottou, Karlen, Kavukcuoglu,
  and Kuksa}]{Collobert:2011jmlr}
Ronan Collobert, Jason Weston, L{\'{e}}on Bottou, Michael Karlen, Koray
  Kavukcuoglu, and Pavel~P. Kuksa. 2011.
\newblock \href{http://dl.acm.org/citation.cfm?id=1953048.2078186}{Natural
  language processing (almost) from scratch}.
\newblock {\em Journal of Machine Learning Research\/} 12:2493--2537.
\newblock
  \href{http://dl.acm.org/citation.cfm?id=1953048.2078186}{http://dl.acm.org/citation.cfm?id=1953048.2078186}.

\bibitem[{Davidov and Rappoport(2006)}]{Davidov:2006acl}
Dmitry Davidov and Ari Rappoport. 2006.
\newblock \href{http://www.aclweb.org/anthology/P06-1038}{Efficient
  unsupervised discovery of word categories using symmetric patterns and high
  frequency words}.
\newblock In {\em ACL\/}. pages 297--304.
\newblock
  \href{http://www.aclweb.org/anthology/P06-1038}{http://www.aclweb.org/anthology/P06-1038}.

\bibitem[{de~Marneffe et~al.(2014)de~Marneffe, Dozat, Silveira, Haverinen,
  Ginter, Nivre, and Manning}]{Marneffe:2014lrec}
Marie{-}Catherine de~Marneffe, Timothy Dozat, Natalia Silveira, Katri
  Haverinen, Filip Ginter, Joakim Nivre, and Christopher~D. Manning. 2014.
\newblock
  \href{http://www.lrec-conf.org/proceedings/lrec2014/summaries/1062.html}{Universal
  {S}tanford dependencies: {A} cross-linguistic typology}.
\newblock In {\em LREC\/}. pages 4585--4592.
\newblock
  \href{http://www.lrec-conf.org/proceedings/lrec2014/summaries/1062.html}{http://www.lrec-conf.org/proceedings/lrec2014/summaries/1062.html}.

\bibitem[{de~Marneffe and Manning(2008)}]{Marneffe:2008sd}
Marie-Catherine de~Marneffe and Christopher~D. Manning. 2008.
\newblock \href{http://www.aclweb.org/anthology/W08-1301}{The {S}tanford typed
  dependencies representation}.
\newblock In {\em Proceedings of the Workshop on Cross-Framework and
  Cross-Domain Parser Evaluation\/}. pages 1--8.
\newblock
  \href{http://www.aclweb.org/anthology/W08-1301}{http://www.aclweb.org/anthology/W08-1301}.

\bibitem[{Faruqui et~al.(2015)Faruqui, Dodge, Jauhar, Dyer, Hovy, and
  Smith}]{Faruqui:2015naacl}
Manaal Faruqui, Jesse Dodge, Sujay~Kumar Jauhar, Chris Dyer, Eduard Hovy, and
  Noah~A. Smith. 2015.
\newblock \href{http://www.aclweb.org/anthology/N15-1184}{Retrofitting word
  vectors to semantic lexicons}.
\newblock In {\em NAACL-HLT\/}. pages 1606--1615.
\newblock
  \href{http://www.aclweb.org/anthology/N15-1184}{http://www.aclweb.org/anthology/N15-1184}.

\bibitem[{Goldberg and Levy(2014)}]{Goldberg:2014arxiv}
Yoav Goldberg and Omer Levy. 2014.
\newblock \href{http://arxiv.org/abs/1402.3722}{Word2vec explained: {D}eriving
  {M}ikolov et al.'s negative-sampling word-embedding method}.
\newblock {\em CoRR\/} abs/1402.3722.
\newblock
  \href{http://arxiv.org/abs/1402.3722}{http://arxiv.org/abs/1402.3722}.

\bibitem[{Goldberg and Nivre(2012)}]{Goldberg:2012coling}
Yoav Goldberg and Joakim Nivre. 2012.
\newblock \href{http://www.aclweb.org/anthology/C12-1059}{A dynamic oracle for
  arc-eager dependency parsing}.
\newblock In {\em COLING\/}. pages 959--976.
\newblock
  \href{http://www.aclweb.org/anthology/C12-1059}{http://www.aclweb.org/anthology/C12-1059}.

\bibitem[{Harris(1954)}]{Harris:1954}
Zellig~S. Harris. 1954.
\newblock \href{https://doi.org/10.1080/00437956.1954.11659520}{Distributional
  structure}.
\newblock {\em Word\/} 10(23):146--162.
\newblock
  \href{https://doi.org/10.1080/00437956.1954.11659520}{https://doi.org/10.1080/00437956.1954.11659520}.

\bibitem[{Kiela et~al.(2015)Kiela, Hill, and Clark}]{Kiela:2015emnlpemb}
Douwe Kiela, Felix Hill, and Stephen Clark. 2015.
\newblock \href{http://aclweb.org/anthology/D15-1242}{Specializing word
  embeddings for similarity or relatedness}.
\newblock In {\em EMNLP\/}. pages 2044--2048.
\newblock
  \href{http://aclweb.org/anthology/D15-1242}{http://aclweb.org/anthology/D15-1242}.

\bibitem[{Landauer and Dumais(1997)}]{Landauer:1997pr}
Thomas~K. Landauer and Susan~T. Dumais. 1997.
\newblock \href{https://doi.org/10.1037/0033-295X.104.2.211}{Solutions to
  {P}lato's problem: {T}he {L}atent {S}emantic {A}nalysis theory of
  acquisition, induction, and representation of knowledge}.
\newblock {\em Psychological Review\/} 104(2):211--240.
\newblock
  \href{https://doi.org/10.1037/0033-295X.104.2.211}{https://doi.org/10.1037/0033-295X.104.2.211}.

\bibitem[{Leviant and Reichart(2015)}]{Leviant:2015arxiv}
Ira Leviant and Roi Reichart. 2015.
\newblock \href{http://arxiv.org/abs/1508.00106}{Separated by an un-common
  language: {T}owards judgment language informed vector space modeling}.
\newblock {\em CoRR\/} abs/1508.00106.
\newblock
  \href{http://arxiv.org/abs/1508.00106}{http://arxiv.org/abs/1508.00106}.

\bibitem[{Levy and Goldberg(2014{\natexlab{a}})}]{Levy:2014acl}
Omer Levy and Yoav Goldberg. 2014{\natexlab{a}}.
\newblock \href{http://www.aclweb.org/anthology/P14-2050}{Dependency-based word
  embeddings}.
\newblock In {\em ACL\/}. pages 302--308.
\newblock
  \href{http://www.aclweb.org/anthology/P14-2050}{http://www.aclweb.org/anthology/P14-2050}.

\bibitem[{Levy and Goldberg(2014{\natexlab{b}})}]{Levy:2014conll}
Omer Levy and Yoav Goldberg. 2014{\natexlab{b}}.
\newblock \href{http://www.aclweb.org/anthology/W14-1618}{Linguistic
  regularities in sparse and explicit word representations}.
\newblock In {\em CoNLL\/}. pages 171--180.
\newblock
  \href{http://www.aclweb.org/anthology/W14-1618}{http://www.aclweb.org/anthology/W14-1618}.

\bibitem[{Levy and Goldberg(2014{\natexlab{c}})}]{Levy:2014nips}
Omer Levy and Yoav Goldberg. 2014{\natexlab{c}}.
\newblock \href{http://papers.nips.cc/paper/5477.pdf}{Neural word embedding as
  implicit matrix factorization}.
\newblock In {\em NIPS\/}. pages 2177--2185.
\newblock
  \href{http://papers.nips.cc/paper/5477.pdf}{http://papers.nips.cc/paper/5477.pdf}.

\bibitem[{Levy et~al.(2015)Levy, Goldberg, and Dagan}]{Levy:2015tacl}
Omer Levy, Yoav Goldberg, and Ido Dagan. 2015.
\newblock Improving distributional similarity with lessons learned from word
  embeddings.
\newblock {\em Transactions of the ACL\/} 3:211--225.

\bibitem[{Ling et~al.(2015{\natexlab{a}})Ling, Dyer, Black, and
  Trancoso}]{Ling:2015naacl}
Wang Ling, Chris Dyer, Alan~W. Black, and Isabel Trancoso. 2015{\natexlab{a}}.
\newblock \href{http://www.aclweb.org/anthology/N15-1142}{Two/too simple
  adaptations of {Word2Vec} for syntax problems}.
\newblock In {\em NAACL-HLT\/}. pages 1299--1304.
\newblock
  \href{http://www.aclweb.org/anthology/N15-1142}{http://www.aclweb.org/anthology/N15-1142}.

\bibitem[{Ling et~al.(2015{\natexlab{b}})Ling, Tsvetkov, Amir, Fermandez, Dyer,
  Black, Trancoso, and Lin}]{Ling:2015emnlp}
Wang Ling, Yulia Tsvetkov, Silvio Amir, Ramon Fermandez, Chris Dyer, Alan~W
  Black, Isabel Trancoso, and Chu-Cheng Lin. 2015{\natexlab{b}}.
\newblock \href{http://aclweb.org/anthology/D15-1161}{Not all contexts are
  created equal: Better word representations with variable attention}.
\newblock In {\em EMNLP\/}. pages 1367--1372.
\newblock
  \href{http://aclweb.org/anthology/D15-1161}{http://aclweb.org/anthology/D15-1161}.

\bibitem[{Liu et~al.(2015)Liu, Jiang, Wei, Ling, and Hu}]{Liu:2015acl}
Quan Liu, Hui Jiang, Si~Wei, Zhen-Hua Ling, and Yu~Hu. 2015.
\newblock \href{http://www.aclweb.org/anthology/P15-1145}{Learning semantic
  word embeddings based on ordinal knowledge constraints}.
\newblock In {\em ACL\/}. pages 1501--1511.
\newblock
  \href{http://www.aclweb.org/anthology/P15-1145}{http://www.aclweb.org/anthology/P15-1145}.

\bibitem[{Martins et~al.(2013)Martins, Almeida, and Smith}]{Martins:2013acl}
Andr{\'{e}} F.~T. Martins, Miguel~B. Almeida, and Noah~A. Smith. 2013.
\newblock \href{http://www.aclweb.org/anthology/P13-2109}{Turning on the
  {Turbo: F}ast third-order non-projective turbo parsers}.
\newblock In {\em ACL\/}. pages 617--622.
\newblock
  \href{http://www.aclweb.org/anthology/P13-2109}{http://www.aclweb.org/anthology/P13-2109}.

\bibitem[{McDonald et~al.(2013)McDonald, Nivre, Quirmbach{-}Brundage, Goldberg,
  Das, Ganchev, Hall, Petrov, Zhang, T{\"{a}}ckstr{\"{o}}m, Bedini,
  Castell{\'{o}}, and Lee}]{McDonald:2013acl}
Ryan~T. McDonald, Joakim Nivre, Yvonne Quirmbach{-}Brundage, Yoav Goldberg,
  Dipanjan Das, Kuzman Ganchev, Keith~B. Hall, Slav Petrov, Hao Zhang, Oscar
  T{\"{a}}ckstr{\"{o}}m, Claudia Bedini, N{\'{u}}ria~Bertomeu Castell{\'{o}},
  and Jungmee Lee. 2013.
\newblock \href{http://www.aclweb.org/anthology/P13-2017}{Universal dependency
  annotation for multilingual parsing}.
\newblock In {\em ACL\/}. pages 92--97.
\newblock
  \href{http://www.aclweb.org/anthology/P13-2017}{http://www.aclweb.org/anthology/P13-2017}.

\bibitem[{Melamud et~al.(2015)Melamud, Dagan, and
  Goldberger}]{Melamud:2015naacl}
Oren Melamud, Ido Dagan, and Jacob Goldberger. 2015.
\newblock \href{http://www.aclweb.org/anthology/N15-1050}{Modeling word meaning
  in context with substitute vectors}.
\newblock In {\em NAACL-HLT\/}. pages 472--482.
\newblock
  \href{http://www.aclweb.org/anthology/N15-1050}{http://www.aclweb.org/anthology/N15-1050}.

\bibitem[{Melamud et~al.(2016)Melamud, McClosky, Patwardhan, and
  Bansal}]{Melamud:2016naacl}
Oren Melamud, David McClosky, Siddharth Patwardhan, and Mohit Bansal. 2016.
\newblock \href{http://www.aclweb.org/anthology/N16-1118}{The role of context
  types and dimensionality in learning word embeddings}.
\newblock In {\em NAACL-HLT\/}.
\newblock
  \href{http://www.aclweb.org/anthology/N16-1118}{http://www.aclweb.org/anthology/N16-1118}.

\bibitem[{Mikolov et~al.(2013)Mikolov, Sutskever, Chen, Corrado, and
  Dean}]{Mikolov:2013nips}
Tomas Mikolov, Ilya Sutskever, Kai Chen, Gregory~S. Corrado, and Jeffrey Dean.
  2013.
\newblock Distributed representations of words and phrases and their
  compositionality.
\newblock In {\em NIPS\/}. pages 3111--3119.

\bibitem[{Mnih and Kavukcuoglu(2013)}]{Mnih:2013nips}
Andriy Mnih and Koray Kavukcuoglu. 2013.
\newblock Learning word embeddings efficiently with noise-contrastive
  estimation.
\newblock In {\em NIPS\/}. pages 2265--2273.

\bibitem[{Mrk\v{s}i\'c et~al.(2017)Mrk\v{s}i\'c, Vuli\'{c}, {\'O S\'eaghdha},
  Leviant, Reichart, Ga\v{s}i\'{c}, Korhonen, and Young}]{Mrksic:2017ar}
Nikola Mrk\v{s}i\'c, Ivan Vuli\'{c}, Diarmuid {\'O S\'eaghdha}, Ira Leviant,
  Roi Reichart, Milica Ga\v{s}i\'{c}, Anna Korhonen, and Steve Young. 2017.
\newblock \href{https://arxiv.org/abs/1706.00374}{Semantic specialisation of
  distributional word vector spaces using monolingual and cross-lingual
  constraints}.
\newblock {\em Transactions of the ACL\/}
  \href{https://arxiv.org/abs/1706.00374}{https://arxiv.org/abs/1706.00374}.

\bibitem[{Nivre~et al.(2016)}]{Nivre:2015ud}
Joakim Nivre~et al. 2016.
\newblock Universal {D}ependencies 1.4.
\newblock {LINDAT}/{CLARIN} digital library at Institute of Formal and Applied
  Linguistics, Charles University in Prague.

\bibitem[{Pad{\'{o}} and Lapata(2007)}]{Pado:2007cl}
Sebastian Pad{\'{o}} and Mirella Lapata. 2007.
\newblock \href{https://doi.org/10.1162/coli.2007.33.2.161}{Dependency-based
  construction of semantic space models}.
\newblock {\em Computational Linguistics\/} 33(2):161--199.
\newblock
  \href{https://doi.org/10.1162/coli.2007.33.2.161}{https://doi.org/10.1162/coli.2007.33.2.161}.

\bibitem[{Pearl(1984)}]{Pearl:1984}
Judea Pearl. 1984.
\newblock Heuristics: {I}ntelligent search strategies for computer problem
  solving .

\bibitem[{Pennington et~al.(2014)Pennington, Socher, and
  Manning}]{Pennington:2014emnlp}
Jeffrey Pennington, Richard Socher, and Christopher Manning. 2014.
\newblock \href{http://www.aclweb.org/anthology/D14-1162}{Glove: Global vectors
  for word representation}.
\newblock In {\em EMNLP\/}. pages 1532--1543.
\newblock
  \href{http://www.aclweb.org/anthology/D14-1162}{http://www.aclweb.org/anthology/D14-1162}.

\bibitem[{Petrov et~al.(2012)Petrov, Das, and McDonald}]{Petrov:2012lrec}
Slav Petrov, Dipanjan Das, and Ryan~T. McDonald. 2012.
\newblock
  \href{http://www.lrec-conf.org/proceedings/lrec2012/summaries/274.html}{A
  universal part-of-speech tagset}.
\newblock In {\em LREC\/}. pages 2089--2096.
\newblock
  \href{http://www.lrec-conf.org/proceedings/lrec2012/summaries/274.html}{http://www.lrec-conf.org/proceedings/lrec2012/summaries/274.html}.

\bibitem[{Sch\"{u}tze(1993)}]{Schutze:1993acl}
Hinrich Sch\"{u}tze. 1993.
\newblock \href{http://www.aclweb.org/anthology/P93-1034}{Part-of-speech
  induction from scratch}.
\newblock In {\em ACL\/}. pages 251--258.
\newblock
  \href{http://www.aclweb.org/anthology/P93-1034}{http://www.aclweb.org/anthology/P93-1034}.

\bibitem[{Schwartz et~al.(2015)Schwartz, Reichart, and
  Rappoport}]{Schwartz:2015conll}
Roy Schwartz, Roi Reichart, and Ari Rappoport. 2015.
\newblock \href{http://www.aclweb.org/anthology/K15-1026}{Symmetric pattern
  based word embeddings for improved word similarity prediction}.
\newblock In {\em CoNLL\/}. pages 258--267.
\newblock
  \href{http://www.aclweb.org/anthology/K15-1026}{http://www.aclweb.org/anthology/K15-1026}.

\bibitem[{Schwartz et~al.(2016)Schwartz, Reichart, and
  Rappoport}]{Schwartz:2016naacl}
Roy Schwartz, Roi Reichart, and Ari Rappoport. 2016.
\newblock \href{http://www.aclweb.org/anthology/N16-1060}{Symmetric patterns
  and coordinations: {F}ast and enhanced representations of verbs and
  adjectives}.
\newblock In {\em NAACL-HLT\/}. pages 499--505.
\newblock
  \href{http://www.aclweb.org/anthology/N16-1060}{http://www.aclweb.org/anthology/N16-1060}.

\bibitem[{Toutanova et~al.(2003)Toutanova, Klein, Manning, and
  Singer}]{Toutanova:2003naacl}
Kristina Toutanova, Dan Klein, Christopher~D. Manning, and Yoram Singer. 2003.
\newblock \href{http://aclweb.org/anthology/N/N03/}{Feature-rich part-of-speech
  tagging with a cyclic dependency network}.
\newblock In {\em NAACL-HLT\/}. pages 173--180.
\newblock
  \href{http://aclweb.org/anthology/N/N03/}{http://aclweb.org/anthology/N/N03/}.

\bibitem[{Turian et~al.(2010)Turian, Ratinov, and Bengio}]{Turian:2010acl}
Joseph~P. Turian, Lev{-}Arie Ratinov, and Yoshua Bengio. 2010.
\newblock \href{http://www.aclweb.org/anthology/P10-1040}{Word representations:
  {A} simple and general method for semi-supervised learning}.
\newblock In {\em ACL\/}. pages 384--394.
\newblock
  \href{http://www.aclweb.org/anthology/P10-1040}{http://www.aclweb.org/anthology/P10-1040}.

\bibitem[{Utt and Pad{\'{o}}(2014)}]{Utt:2014tacl}
Jason Utt and Sebastian Pad{\'{o}}. 2014.
\newblock Crosslingual and multilingual construction of syntax-based vector
  space models.
\newblock {\em Transactions of the ACL\/} 2:245--258.

\bibitem[{Vuli\'{c}(2017)}]{Vulic:2017eacl}
Ivan Vuli\'{c}. 2017.
\newblock \href{http://www.aclweb.org/anthology/E17-2065}{Cross-lingual
  syntactically informed distributed word representations}.
\newblock In {\em EACL\/}. pages 408--414.
\newblock
  \href{http://www.aclweb.org/anthology/E17-2065}{http://www.aclweb.org/anthology/E17-2065}.

\bibitem[{Vuli\'{c} and Korhonen(2016)}]{Vulic:2016acluniversal}
Ivan Vuli\'{c} and Anna Korhonen. 2016.
\newblock \href{http://anthology.aclweb.org/P16-2084}{Is ``universal syntax''
  universally useful for learning distributed word representations?}
\newblock In {\em ACL\/}. pages 518--524.
\newblock
  \href{http://anthology.aclweb.org/P16-2084}{http://anthology.aclweb.org/P16-2084}.

\bibitem[{Wieting et~al.(2015)Wieting, Bansal, Gimpel, and
  Livescu}]{Wieting:2015tacl}
John Wieting, Mohit Bansal, Kevin Gimpel, and Karen Livescu. 2015.
\newblock \href{http://aclweb.org/anthology/Q15-1025}{From paraphrase database
  to compositional paraphrase model and back}.
\newblock {\em Transactions of the ACL\/} 3:345--358.
\newblock
  \href{http://aclweb.org/anthology/Q15-1025}{http://aclweb.org/anthology/Q15-1025}.

\bibitem[{Yatbaz et~al.(2012)Yatbaz, Sert, and Yuret}]{Yatbaz:2012emnlp}
Mehmet~Ali Yatbaz, Enis Sert, and Deniz Yuret. 2012.
\newblock \href{http://www.aclweb.org/anthology/D12-1086}{Learning syntactic
  categories using paradigmatic representations of word context}.
\newblock In {\em EMNLP\/}. pages 940--951.
\newblock
  \href{http://www.aclweb.org/anthology/D12-1086}{http://www.aclweb.org/anthology/D12-1086}.

\bibitem[{Yu and Dredze(2014)}]{Yu:2014acl}
Mo~Yu and Mark Dredze. 2014.
\newblock \href{http://www.aclweb.org/anthology/P14-2089}{Improving lexical
  embeddings with semantic knowledge}.
\newblock In {\em ACL\/}. pages 545--550.
\newblock
  \href{http://www.aclweb.org/anthology/P14-2089}{http://www.aclweb.org/anthology/P14-2089}.

\end{thebibliography}
\bibliographystyle{acl_natbib}

\end{document}